# Attention in Natural Language Processing

Andrea Galassi, Marco Lippi, and Paolo Torroni

***Abstract*—Attention is an increasingly popular mechanism used in a wide range of neural architectures. The mechanism itself has been realized in a variety of formats. However, because of the fast-paced advances in this domain, a systematic overview of attention is still missing. In this article, we define a unified model for attention architectures in natural language processing, with a focus on those designed to work with vector representations of the textual data. We propose a taxonomy of attention models according to four dimensions: the representation of the input, the compatibility function, the distribution function, and the multiplicity of the input and/or output. We present the examples of how prior information can be exploited in attention models and discuss ongoing research efforts and open challenges in the area, providing the first extensive categorization of the vast body of literature in this exciting domain.**

## I. Introduction

IN MANY problems that involve the processing of natural language, the elements composing the source text are characterized by having each a different relevance to the task at hand. For instance, in aspect-based sentiment analysis, cue words, such as "good" or "bad," could be relevant to some aspects under consideration, but not to others. In machine translation, some words in the source text could be irrelevant in the translation of the next word. In a visual question-answering task, background pixels could be irrelevant in answering a question regarding an object in the foreground but relevant to questions regarding the scenery.

Arguably, effective solutions to such problems should factor in a notion of relevance, so as to focus the computational resources on a restricted set of important elements. One possible approach would be to tailor solutions to the specific genre at hand, in order to better exploit known regularities of the input, by feature engineering. For example, in the argumentative analysis of persuasive essays, one could decide to give special emphasis to the final sentence. However, such an approach is not always viable, especially if the input is long or very information-rich, such as in text summarization, where the output is the condensed version of a possibly lengthy text sequence. Another approach of increasing popularity amounts to machine learning the relevance of input elements. In that way, neural architectures could automatically weigh the relevance of any region of the input and take such a weight into account while performing the main task. The commonest solution to this problem is a mechanism known as attention.

Attention was first introduced in natural language processing (NLP) for machine translation tasks by Bahdanau *et al.* [2]. However, the idea of glimpses had already been proposed in computer vision by Larochelle and Hinton [3], following the observation that biological retinas fixate on relevant parts of the optic array, while resolution falls off rapidly with eccentricity. The term visual attention became especially popular after Mnih *et al.* [4] significantly outperformed the state of the art in several image classification tasks as well as in dynamic visual control problems such as object tracking due to an architecture that could adaptively select and then process a sequence of regions or locations at high resolution and use a progressively lower resolution for further pixels.

Besides offering a performance gain, the attention mechanism can also be used as a tool for interpreting the behavior of neural architectures, which are notoriously difficult to understand. Indeed, neural networks are subsymbolic architectures; therefore, the knowledge they gather is stored in numeric elements that do not provide any means of interpretation by themselves. It then becomes hard if not impossible to pinpoint the reasons behind the wrong output of a neural architecture. Interestingly, attention could provide a key to partially interpret and explain neural network behavior [5]–[9], even if it cannot be considered a reliable means of explanation [10], [11]. For instance, the weights computed by attention could point us to relevant information discarded by the neural network or to irrelevant elements of the input source that have been factored in and could explain a surprising output of the neural network.

Therefore, visual highlights of attention weights could be instrumental in analyzing the outcome of neural networks, and a number of specific tools have been devised for such a visualization [12], [13]. Fig. 1 shows an example of attention visualization in the context of aspect-based sentiment analysis.

For all these reasons, attention has become an increasingly common ingredient of neural architectures for NLP [14], [15]. Table I presents a nonexhaustive list of neural architectures where the introduction of an attention mechanism has brought about a significant gain. Works are grouped by the NLP tasks they address. The spectrum of tasks involved is remarkably broad. Besides NLP and computer vision [16]–[18], attentive models have been successfully adopted in many other different fields, such as speech recognition [19]–[21],

Manuscript received July 5, 2019; revised December 17, 2019 and April 17, 2020; accepted August 20, 2020. Date of publication September 10, 2020; date of current version October 6, 2021. This work was supported by Horizon 2020, project AI4EU, under Grant 825619. *(Corresponding author: Andrea Galassi.)*

Andrea Galassi and Paolo Torroni are with the Department of Computer Science and Engineering (DISI), University of Bologna, 40126 Bologna, Italy (e-mail: a.galassi@unibo.it; paolo.torroni@unibo.it).

Marco Lippi is with the Department of Sciences and Methods for Engineering (DISMI), University of Modena and Reggio Emilia, 41121 Modena, Italy (e-mail: marco.lippi@unimore.it).

Color versions of one or more of the figures in this article are available online at https://ieeexplore.ieee.org.

*Task: Hotel location*

you get what you pay for . not the cleanest rooms but bed was clean and so was bathroom . bring your own towels though as very thin . service was excellent , let us book in at 8:30am ! **for location and price , this ca n't be beaten** , but it is cheap for a reason . if you come expecting the hilton , then book the hilton ! for uk travellers , think of a blackpool b&b.

*Task: Hotel cleanliness*

you get what you pay for . **not the cleanest rooms but bed was clean and so was bathroom** . bring your own towels though as very thin . service was excellent , let us book in at 8:30am ! for location and price , this ca n't be beaten , but it is cheap for a reason . if you come expecting the hilton , then book the hilton ! for uk travellers , think of a blackpool b&b.

*Task: Hotel service*

you get what you pay for . not the cleanest rooms but bed was clean and so was bathroom . bring your own towels though as very thin . **service was excellent** , let us book in at 8:30am ! for location and price , this ca n't be beaten , but it is cheap for a reason . if you come expecting the hilton , then book the hilton ! for uk travellers , think of a blackpool b&b.

Fig. 1. Example of attention visualization for an aspect-based sentiment analysis task, from [1, Fig. 6]. Words are highlighted according to attention scores. Phrases in bold are the words considered relevant for the task or human rationales.

TABLE I

NONEXHAUSTIVE LIST OF WORKS THAT EXPLOIT ATTENTION, GROUPED BY THE TASK(S) ADDRESSED

| Task Addressed | Related Works |
|---|---|
| **Machine Translation** | [2, 6, 8, 29–48, 48–50] |
| Translation Quality Estimation | [51] |
| **Text Classification** | [7, 8, 10, 11, 52, 53] |
| Abusive content detection | [54] |
| **Text Summarization** | [41, 55–58] |
| **Language Modelling** | [59–61] |
| **Question Answering** | [10, 47, 59, 62–75] |
| Question Answering over Knowledge Base | [76] |
| **Morphology** | |
| Pun Recognition | [77] |
| **Multimodal Tasks** | [78] |
| Image Captioning | [16, 79] |
| Visual Question Answering | [80–82] |
| Task-oriented Language Grounding | [83] |
| **Information Extraction** | |
| Coreference Resolution | [84, 85] |
| Named Entity Recognition | [51, 86] |
| Optical Character Recognition Correction | [87] |
| **Semantic** | |
| Entity Disambiguation | [88] |
| Natural Language Inference | [8, 10, 47, 89–96] |
| Semantic Relatedness | [93] |
| Semantic Role Labelling | [97, 98] |
| Sentence Similarity | [96] |
| Textual Entailment | [75, 99, 100] |
| Word Sense Disambiguation | [101] |
| **Syntax** | |
| Constituency Parsing | [102, 103] |
| Dependency Parsing | [51, 104, 105] |
| **Sentiment Analysis** | [1, 7, 93, 95, 100, 106–120] |
| Agreement/Disagreement Identification | [121] |
| Argumentation Mining | [57, 122–125] |
| Emoji prediction | [126] |
| Emotion Cause Analysis | [127, 128] |
| Emotion Classification | [115] |

recommendation [22], [23], time-series analysis [24], [25], games [26], and mathematical problems [27], [28].

In NLP, after an initial exploration by a number of seminal papers [2], [59], a fast-paced development of new attention models and attentive architectures ensued, resulting in a highly diversified architectural landscape. Because of, and adding to, the overall complexity, it is not unheard of different authors who have been independently following similar intuitions leading to the development of almost identical attention models. For instance, the concepts of inner attention [68] and word attention [41] are arguably one and the same. Unsurprisingly, the same terms have been introduced by different authors to define different concepts, thus further adding to the ambiguity in the literature. For example, the term context vector is used with different meanings by Bahdanau *et al.* [2], Yang *et al.* [52], and Wang *et al.* [129].

In this article, we offer a systematic overview of attention models developed for NLP. To this end, we provide a general model of attention for NLP tasks and use it to chart the major research activities in this area. We also introduce a taxonomy that describes the existing approaches along four dimensions: input representation, compatibility function, distribution function, and input/output multiplicity. To the best of our knowledge, this is the first taxonomy of attention models. Accordingly, we provide a succinct description of each attention model, compare the models with one another, and offer insights on their use. Moreover, we present the examples regarding the use of prior information in unison with attention, debate about the possible future uses of attention, and describe some interesting open challenges.

We restrict our analysis to attentive architectures designed to work with vector representation of data, as it typically is the case in NLP. Readers interested in attention models for tasks where data have a graphical representation may refer to Lee *et al.* [130].

What this survey does not offer is a comprehensive account of all the neural architectures for NLP (for an excellent overview, see [131]) or of all the neural architectures for NLP that uses an attention mechanism. This would be impossible and would rapidly become obsolete because of the sheer volume of new articles featuring architectures that increasingly rely on such a mechanism. Moreover, our purpose is to produce a synthesis and a critical outlook rather than a flat listing of research activities. For the same reason, we do not offer a quantitative evaluation of different types of attention mechanisms since such mechanisms are generally embedded in larger neural network architectures devised to address specific tasks, and it would be pointless in many cases to attempt comparisons using different standards. Even for a single specific NLP task, a fair evaluation of different attention models would require experimentation with multiple neural architectures, extensive hyperparameter tuning, and validation over a variety of benchmarks. However, attention can be applied to a multiplicity of tasks, and there are no data sets that would meaningfully cover such a variety of tasks. An empirical evaluation is thus beyond the scope of this article. There are, however, a number of experimental studies focused on particular NLP tasks, including machine translation [37], [42], [48], [132], argumentation mining [125], text summarization [58], and sentiment analysis [7]. It is worthwhile remarking that, on several occasions, attention-based approaches enabled a dramatic development of entire research lines. In some cases, such a development has produced an immediate performance boost. This was the case, for example, with the transformer [36] for sequence-to-sequence annotation, as well as with BERT [60], currently among the most popular architectures for the creation of embeddings. In other cases, the impact of attention-based models was even greater, paving the way to radically new approaches for some tasks. This was the influence of Bahdanau *et al.*'s work [2] to the field of machine translation. Likewise, the expressive power of memory networks [59] significantly contributed to the idea of using deep networks for reasoning tasks.

This survey is structured as follows. In Section II, we define a general model of attention and describe its components. We use a well-known machine-translation architecture introduced by Bahdanau *et al.* [2] as an illustration and an instance of the general model. In Section III, we elaborate on the uses of attention in various NLP tasks. Section IV presents our taxonomy of attention models. Section V discusses how attention can be combined with knowledge about the task or the data. Section VI is devoted to open challenges, current trends, and future directions. Section VII concludes this article.

## II. Attention Function

The attention mechanism is a part of a neural architecture that enables to dynamically highlight relevant features of the input data, which, in NLP, is typically a sequence of textual elements. It can be applied directly to the raw input or to its higher level representation. The core idea behind attention is to compute a weight distribution on the input sequence, assigning higher values to more relevant elements.

To illustrate, we briefly describe a classic attention architecture, called RNNsearch [2]. We chose RNNsearch because of its historical significance and for its simplicity with respect to other structures that we will describe further on.

### A. *Example for Machine Translation and Alignment*

RNNsearch uses attention for machine translation. The objective is to compute an output sequence $\boldsymbol{y}$ that is a translation of an input sequence $\boldsymbol{x}$. The architecture consists of an encoder followed by a decoder, as shown in Fig. 2.

The encoder is a bidirectional recurrent neural network (BiRNN) [133] that computes an annotation term $\boldsymbol{h_i}$ for every

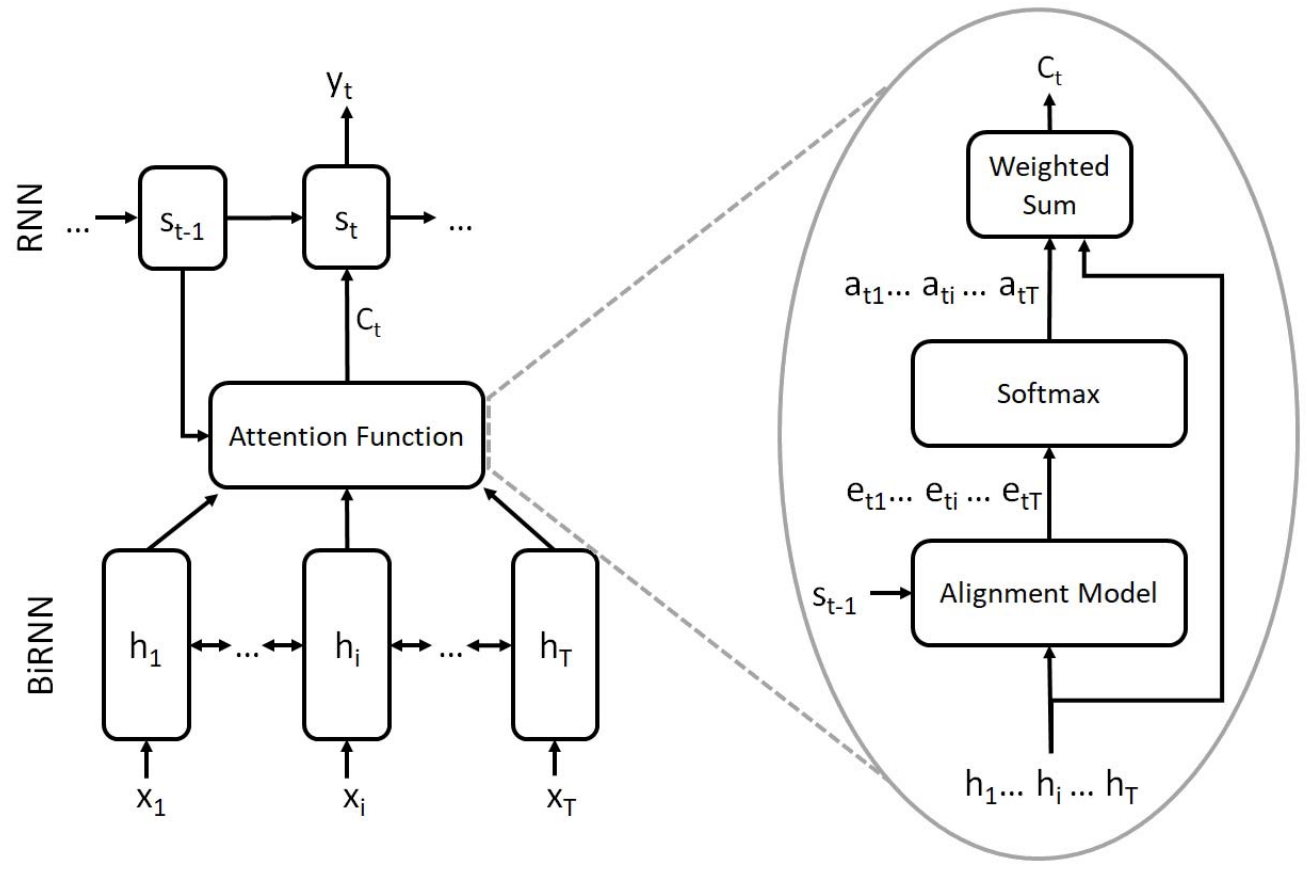

Fig. 2. Architecture of RNNsearch [2] (left) and its attention model (right).

input term $x_i$ of $\boldsymbol{x}$

$$(\boldsymbol{h_1}, \ldots, \boldsymbol{h_T}) = \text{BiRNN}(x_1, \ldots, x_T). \tag{1}$$

The decoder consists of two cascading elements: the attention function and an RNN. At each time step $t$, the attention function produces an embedding $\boldsymbol{c_t}$ of the input sequence, called a context vector. The subsequent RNN, characterized by a hidden state $\boldsymbol{s_t}$, computes from such an embedding a probability distribution over all possible output symbols, pointing to the most probable symbol $y_t$

$$p(y_t | y_1, \ldots, y_{t-1}, \boldsymbol{x}) = \text{RNN}(\boldsymbol{c_t}). \tag{2}$$

The context vector is obtained as follows. At each time step $t$, the attention function takes as input the previous hidden state of the RNN $s_{t-1}$ and the annotations $\boldsymbol{h_1}, \ldots, \boldsymbol{h_T}$. Such inputs are processed through an alignment model [see (3)] to obtain a set of scalar values $e_{ti}$ that score the matching between the inputs around position $i$ and the outputs around position $t$. These scores are then normalized through a softmax function, so as to obtain a set of weights $a_{ti}$ [see (4)]

$$e_{ti} = f(\boldsymbol{s_{t-1}}, \boldsymbol{h_i}) \tag{3}$$

$$a_{ti} = \frac{\exp(e_{ti})}{\sum_{j=1}^{T} \exp(e_{tj})}. \tag{4}$$

Finally, the context vector $\boldsymbol{c_t}$ is computed as a weighted sum of the annotations $\boldsymbol{h_i}$ based on their weights $a_{ti}$

$$\boldsymbol{c_t} = \sum_i a_{ti} \boldsymbol{h_i}. \tag{5}$$

Quoting Bahdanau *et al.* [2], the use of attention "relieve[s] the encoder from the burden of having to encode all information in the source sentence into a fixed-length vector. With this new approach, the information can be spread throughout the sequence of annotations, which can be selectively retrieved by the decoder accordingly."

### B. *Unified Attention Model*

The characteristics of an attention model depend on the structure of the data whereupon they operate and on the desired output structure. The unified model we propose is based on and extends the models proposed by Vaswani *et al.* [36] and Daniluk *et al.* [61]. It comprises a core part shared by almost the totality of the models found in the surveyed literature, as well as some additional components that, although not universally present, are still found in most literature models.

TABLE II
NOTATION

| Symbol | Name | Definition |
|---|---|---|
| $\boldsymbol{x}$ | Input sequence | Sequence of textual elements constituting the raw input. |
| $\boldsymbol{K}$ | Keys | Matrix of $d_k$ vectors ($\boldsymbol{k_i}$) of size $n_k$, whereupon attention weights are computed: $\boldsymbol{K} \in \mathbb{R}^{n_k \times d_k}$. |
| $\boldsymbol{V}$ | Values | Matrix of $d_k$ vectors ($\boldsymbol{v_i}$) of size $n_v$, whereupon attention is applied: $\boldsymbol{V} \in \mathbb{R}^{n_v \times d_k}$. Each $\boldsymbol{v_i}$ and its corresponding $\boldsymbol{k_i}$ offer two, possibly different, interpretations of the same entity. |
| $\boldsymbol{q}$ | Query | Vector of size $n_q$, or sequence thereof, in which respect attention is computed: $\boldsymbol{q} \in \mathbb{R}^{n_q}$. |
| *kaf* *qaf* *vaf* | Annotation functions | Functions that encode the input sequence and query, producing $\boldsymbol{K}$, $\boldsymbol{q}$ and $\boldsymbol{V}$ respectively. |
| $\boldsymbol{e}$ | Energy scores | Vector of size $d_k$, whose scalar elements (energy "scores", $e_i$) represent the relevance of the corresponding $\boldsymbol{k_i}$, according to the compatibility function: $\boldsymbol{e} \in \mathbb{R}^{d_k}$. |
| $\boldsymbol{a}$ | Attention weights | Vector of size $d_k$, whose scalar elements (attention "weights", $a_i$) represent the relevance of the corresponding $\boldsymbol{k_i}$ according to the attention model: $\boldsymbol{a} \in \mathbb{R}^{d_k}$. |
| $f$ | Compatibility function | Function that evaluates the relevance of $\boldsymbol{K}$ with respect to $\boldsymbol{q}$, returning a vector of energy scores: $\boldsymbol{e} = f(\boldsymbol{K}, \boldsymbol{q})$. |
| $g$ | Distribution function | Function that computes the attention weights from the energy scores: $\boldsymbol{a} = g(\boldsymbol{e})$. |
| $\boldsymbol{Z}$ | Weighted values | Matrix of $d_k$ vectors ($\boldsymbol{z_i}$) of size $n_v$, representing the application of $\boldsymbol{a}$ to $\boldsymbol{V}$: $\boldsymbol{Z} \in \mathbb{R}^{n_v \times d_k}$. |
| $\boldsymbol{c}$ | Context vector | Vector of size $n_v$, offering a compact representation of $\boldsymbol{Z}$: $\boldsymbol{c} \in \mathbb{R}^{n_v}$. |

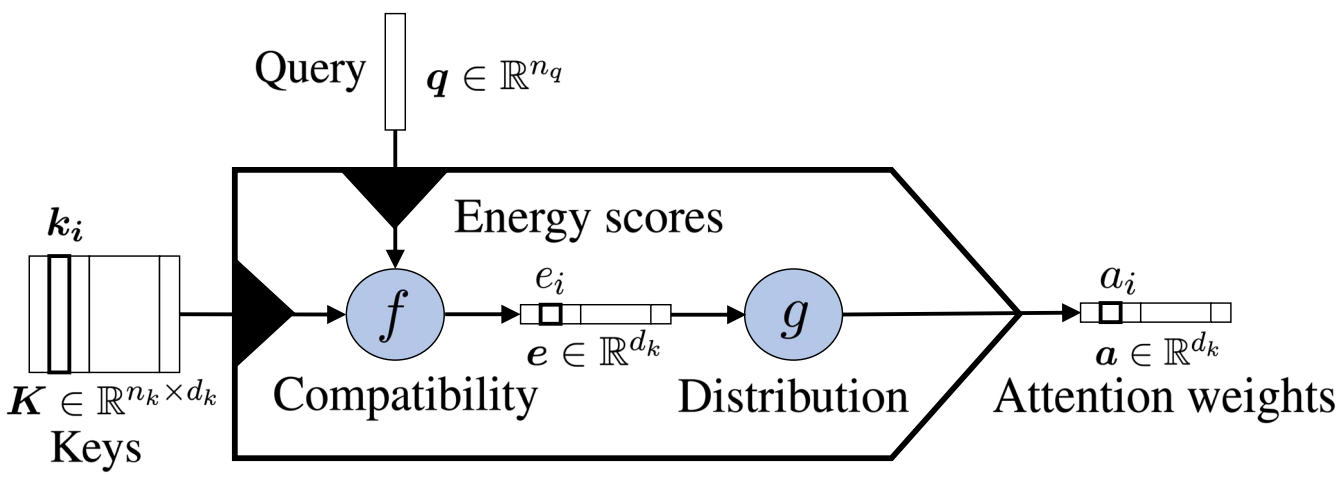

Fig. 3. Core attention model.

Fig. 3 shows the core attention model, which is part of the general model shown in Fig. 4. Table II lists the key terms and symbols. The core of the attention mechanism maps a sequence $\boldsymbol{K}$ of $d_k$ vectors $\boldsymbol{k_i}$, the keys, to a distribution $\boldsymbol{a}$ of $d_k$ weights $a_i$. $\boldsymbol{K}$ encodes the data features whereupon attention is computed. For instance, $\boldsymbol{K}$ may be word or character embeddings of a document, or the internal states of a recurrent architecture, as it happens with the annotation $\boldsymbol{h_i}$ in RNNsearch. In some cases, $\boldsymbol{K}$ could include multiple features or representations of the same object (e.g., both one-hot encoding and embedding of a word) or even—if the task calls for it—representations of entire documents.

More often than not, another input element $\boldsymbol{q}$, called query,[1] is used as a reference when computing the attention distribution. In that case, the attention mechanism will give emphasis to the input elements relevant to the task according to $\boldsymbol{q}$. If no query is defined, attention will give emphasis to the elements inherently relevant to the task at hand. In RNNsearch, for instance, $\boldsymbol{q}$ is a single element, namely, the RNN hidden state $\boldsymbol{s_{t-1}}$. In other architectures, $\boldsymbol{q}$ may represent different entities: embeddings of actual textual queries, contextual information, background knowledge, and so on. It can also take the form of a matrix rather than a vector. For example, in their document attentive reader, Sordoni *et al.* [67] made use of two query vectors.

From the keys and query, a vector $\boldsymbol{e}$ of $d_k$ energy scores $e_i$ is computed through a compatibility function $f$

$$\boldsymbol{e} = f(\boldsymbol{q}, \boldsymbol{K}). \tag{6}$$

Function $f$ corresponds to RNNsearch's alignment model and to what other authors call energy function [43]. Energy scores are then transformed into attention weights using what we call a distribution function $g$

$$\boldsymbol{a} = g(\boldsymbol{e}). \tag{7}$$

Such weights are the outcome of the core attention mechanism. The commonest distribution function is the softmax function, as in RNNsearch, which normalizes all the scores to a probability distribution. Weights represent the relevance of each element to the given task, with respect to $\boldsymbol{q}$ and $\boldsymbol{K}$.

The computation of these weights may already be sufficient for some tasks, such as the classification task addressed by Cui *et al.* [70]. Nevertheless, many tasks require the computation of new representation of the keys. In such cases, it is common to have another input element; a sequence $\boldsymbol{V}$ of $d_k$ vectors $\boldsymbol{v_i}$, the values, representing the data whereupon the attention computed from $\boldsymbol{K}$ and $\boldsymbol{q}$ is to be applied. Each element of $\boldsymbol{V}$ corresponds to one and only one element of $\boldsymbol{K}$, and the two can be seen as different representations of the same data. Indeed, many architectures, including RNNsearch, do not distinguish between $\boldsymbol{K}$ and $\boldsymbol{V}$. The distinction between keys and values was introduced by Daniluk *et al.* [61], who use different representations of the input for computing the attention distribution and the contextual information.

$\boldsymbol{V}$ and $\boldsymbol{a}$ are thus combined to obtain a new set $\boldsymbol{Z}$ of weighted representations of $\boldsymbol{V}$ [see (8)], which are then merged together so as to produce a compact representation of $\boldsymbol{Z}$ usually called the context vector $\boldsymbol{c}$ [see (9)].[2] The commonest way of obtaining $\boldsymbol{c}$ from $\boldsymbol{Z}$ is by summation. However, alternatives have been proposed, including gating functions [93]. Nevertheless, $\boldsymbol{c}$ will be mainly determined by values associated with higher attention weights

$$\boldsymbol{z_i} = a_i \boldsymbol{v_i} \tag{8}$$

$$\boldsymbol{c} = \sum_{i=1}^{d_k} \boldsymbol{z_i}. \tag{9}$$

[1]The concept of "query" in attention models should not be confused with that used in tasks like question answering or information retrieval. In our model, the "query" is part of a general architecture and is task-independent.

[2]Although most authors use this terminology, we shall remark that Yang *et al.* [52], Wang *et al.* [129], and other authors used the term context vector to refer to other elements of the attention architecture.

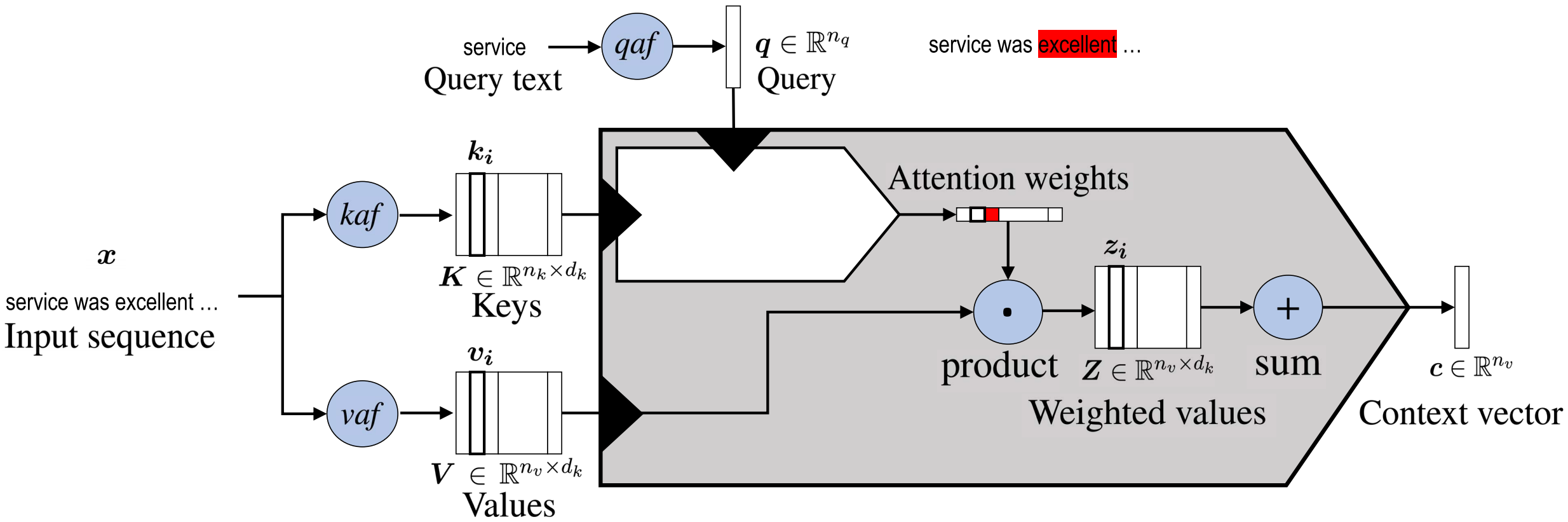

Fig. 4. General attention model.

What we described so far was a synthesis of the most frequent architectural choices made in the design of attentive architectures. Other options will be explored in Section IV-D.

### *C. Deterministic Versus Probabilistic Attention*

Before we proceed, a brief remark about some relevant naming conventions is in order. The attention model described so far is sometimes described in the literature as a mapping with a probabilistic interpretation since the use of a softmax normalization allows one to interpret the attention weights as a probability distribution function (see [91]). Accordingly, some recent literature defines as deterministic attention models [134], [135] and those models where context words, whereupon attention is focused, are deterministically selected, for example, by using the constituency parse tree of the input sentence. However, other authors described the first model as deterministic (soft) attention, to contrast it with stochastic (hard) attention, where the probability distribution over weights is used to hardly sample a single input as context vector $\boldsymbol{c}$, in particular following a reinforcement learning strategy [16]. If the weights $\boldsymbol{a}$ encode a probability distribution, the stochastic attention model differs from our general model for the absence of (8) and (9) that are replaced by the following equations:

$$\tilde{s} \sim \text{Multinoulli}(\{a_i\}) \tag{10}$$

$$\boldsymbol{c} = \sum_{i=1}^{d_k} \tilde{s}_i \boldsymbol{v_i} \quad \text{with } \tilde{s}_i \in \{0, 1\}. \tag{11}$$

To avoid any potential confusion, in the remainder of this article, we will abstain from characterizing the attention models as deterministic, probabilistic, or stochastic.

TABLE III

POSSIBLE USES OF ATTENTION AND EXAMPLES OF RELEVANT TASK

| Use | Tasks |
| --- | --- |
| Feature selection | Multimodal tasks |
| Auxiliary task | Visual question answering<br>Semantic role labelling |
| Contextual embedding creation | Machine Translation<br>Sentiment Analysis<br>Information Extraction |
| Sequence-to-sequence annotation | Machine Translation |
| Word selection | Dependency Parsing<br>Cloze Question Answering |
| Multiple input processing | Question Answering |

## III. USES OF ATTENTION

Attention enables us to estimate the relevance of the input elements as well as to combine said elements into a compact representation—the context vector—that condenses the characteristics of the most relevant elements. Because the context vector is smaller than the original input, it requires fewer computational resources to be processed at later stages, yielding a computational gain.

We summarize possible uses of attention and the tasks in which they are relevant in Table III.

For tasks such as document classification, where usually there is only $\boldsymbol{K}$ in input and no query, the attention mechanism can be seen as an instrument to encode the input into a compact form. The computation of such an embedding can be seen as a form of feature selection, and as such, it can be applied to any set of features sharing the same representation. This applies to cases where features come from different domains as in multimodal tasks [78] or from different levels of a neural architecture [38] or where they simply represent different aspects of the input document [136]. Similarly, attention can also be exploited as an auxiliary task during training so that specific features can be modeled via a multitask setting. This holds for several scenarios, such as visual question answering [137], domain classification for natural language understanding [138], and semantic role labeling [97].

When the generation of a text sequence is required, as in machine translation, attention enables us to make use of a dynamic representation of the input sequence, whereby the whole input does not have to be encoded into a single vector. At each time step, the encoding is tailored according to the task, and in particular, $\boldsymbol{q}$ represents an embedding of the previous state of the decoder. More generally, the possibility to perform attention with respect to a query $\boldsymbol{q}$ allows us to create representations of the input that depend on the task *context*, creating specialized embeddings. This is particularly useful in tasks, such as sentiment analysis and information extraction.

Since attention can create contextual representations of an element, it can also be used to build sequence-to-sequence annotators, without resorting to RNNs or convolutional neural networks (CNNs), as suggested by Vaswani *et al.* [36], who rely on an attention mechanism to obtain a whole encoder/decoder architecture.

Attention can also be used as a tool for selecting specific words. This could be the case, for example, in dependence parsing [97] and in cloze question-answering tasks [66], [70]. In the former case, attention can be applied to a sentence in order to predict dependences. In the latter, attention can be applied to a textual document or to a vocabulary to perform a classification among the words.

Finally, attention can come in handy when multiple interacting input sequences have to be considered in combination. In tasks such as question answering, where the input consists of two textual sequences—for instance, the question and the document or the question and the possible answers—an input encoding can be obtained by considering the mutual interactions between the elements of such sequences, rather than by applying a more rigid *a priori* defined model.

## IV. Taxonomy for Attention Models

Attention models can be described on the basis of the following orthogonal dimensions: the nature of inputs (see Section IV-A), the compatibility function (see Section IV-B), the distribution function (see Section IV-C), and the number of distinct inputs/outputs, which we refer to as "multiplicity" (see Section IV-D). Moreover, attention modules can themselves be used inside larger attention models to obtain complex architectures such as hierarchical-input models (see Section IV-A2) or in some multiple-input coattention models (see Section IV-D2).

### A. Input Representation

In NLP-related tasks, generally, $\boldsymbol{K}$ and $\boldsymbol{V}$ are representations of parts of documents, such as sequences of characters, words, or sentences. These components are usually embedded into continuous vector representations and then processed through key/value annotation functions (called kaf and vaf in Fig. 4), so as to obtain a hidden representation resulting in $\boldsymbol{K}$ and $\boldsymbol{V}$. Typical annotation functions are RNN layers such as gated recurrent units (GRUs), long short-term memory networks (LSTMs), and CNNs. In this way, $\boldsymbol{k_i}$ and $\boldsymbol{v_i}$ represent an input element relative to its local context. If the layers in charge of annotation are trained together with the attention model, they can learn to encode information useful to the attention model.

Alternatively, $\boldsymbol{k_i}/\boldsymbol{v_i}$ can be taken to represent each input element in isolation, rather than in context. For instance, they could be a one-hot encoding of words or characters or a pretrained word embedding. This results in an application of the attention mechanism directly to the raw inputs, which is a model known as inner attention [68]. Such a model has proven to be effective by several authors, who have exploited it in different fashions [36], [41], [54], [116]. The resulting architecture has a smaller number of layers and hyperparameters, which reduces the computational resources needed for training.

$\boldsymbol{K}$ can also represent a single element of the input sequence. This is the case, for example, in the work by Bapna *et al.* [38], whose attention architecture operates on different encodings of the same element, obtained by the subsequent application of RNN layers. The context embeddings obtained for all components individually can then be concatenated, producing a new representation of the document that encodes the most relevant representation of each component for the given task. It would also be possible to aggregate each key or value with its neighbors, by computing their average or sum [128].

We have so far considered the input to be a sequence of characters, words, or sentences, which is usually the case. However, the input can also be other things, such as a juxtaposition of features or relevant aspects of the same textual element. For instance, Li *et al.* [56] and Zadeh *et al.* [78] considered the inputs composed of different sources, and in [136] and [139], the input represents different aspects of the same document. In that case, embeddings of the input can be collated together and fed into an attention model as multiple keys, as long as the embeddings share the same representation. This allows us to highlight the most relevant elements of the inputs and operate a feature selection, leading to a possible reduction of the dimensionality of the representation via the context embedding $\boldsymbol{c}$ produced by the attention mechanism. Interestingly, Li *et al.* [120] proposed a model in which the textual input sequence is mixed with the output of the attention model. Their truncated history-attention model iterates the computation of attention on top of a bi-LSTM. At each step, the bi-LSTM hidden states are used as keys and values, while the context vectors computed in the previous steps act as a query.

We shall now explain in more detail two successful structures, which have become well-established building blocks of neural approaches for NLP, namely, self-attention and hierarchical-input architectures.

*1) Self-Attention:* We made a distinction between two input sources: the input sequence, represented by $\boldsymbol{K}$ and $\boldsymbol{V}$, and the query, represented by $\boldsymbol{q}$. However, some architectures compute attention only based on the input sequence. These architectures are known as self-attentive or intraattentive models. We shall remark, however, that these terms are used to indicate many different approaches. The commonest one amounts to the application of multiple steps of attention to a vector $\boldsymbol{K}$, using the elements $\boldsymbol{k_t}$ of the same vector as query at each step [18], [36]. At each step, the weights $a_i^t$ represent the relevance of $\boldsymbol{k_i}$ with respect to $\boldsymbol{k_t}$, yielding $d_K$ separate context embeddings, $\boldsymbol{c^t}$, one per key. Attention could thus be used as a sequence-to-sequence model, as an alternative to CNNs or RNNs (see Fig. 5). In this way, each element of the new sequence may be influenced by elements of the whole input, incorporating contextual information without any locality boundaries. This is especially interesting since it could overcome a well-known shortcoming of RNNs: their limited ability of modeling long-range dependences [140]. For each element $\boldsymbol{k_t}$, the resulting distribution of the weights $\boldsymbol{a^t}$ should give more emphasis to words that strongly relate to $\boldsymbol{k_t}$. The analysis of these distributions will, therefore, provide information regarding the relationship between the elements inside the sequence. Modern text-sequence generation systems often rely on this approach. Another possibility is to construct a single query element $\boldsymbol{q}$ from the keys through a pooling operation. Furthermore, the same input sequence could be used both as keys $\boldsymbol{K}$ and query $\boldsymbol{Q}$, applying a technique we will describe in Section IV-D2, known as coattention. Other self-attentive approaches, such as [52] and [100], are characterized by the complete absence of any query term $\boldsymbol{q}$, which results in simplified compatibility functions (see Section IV-B).

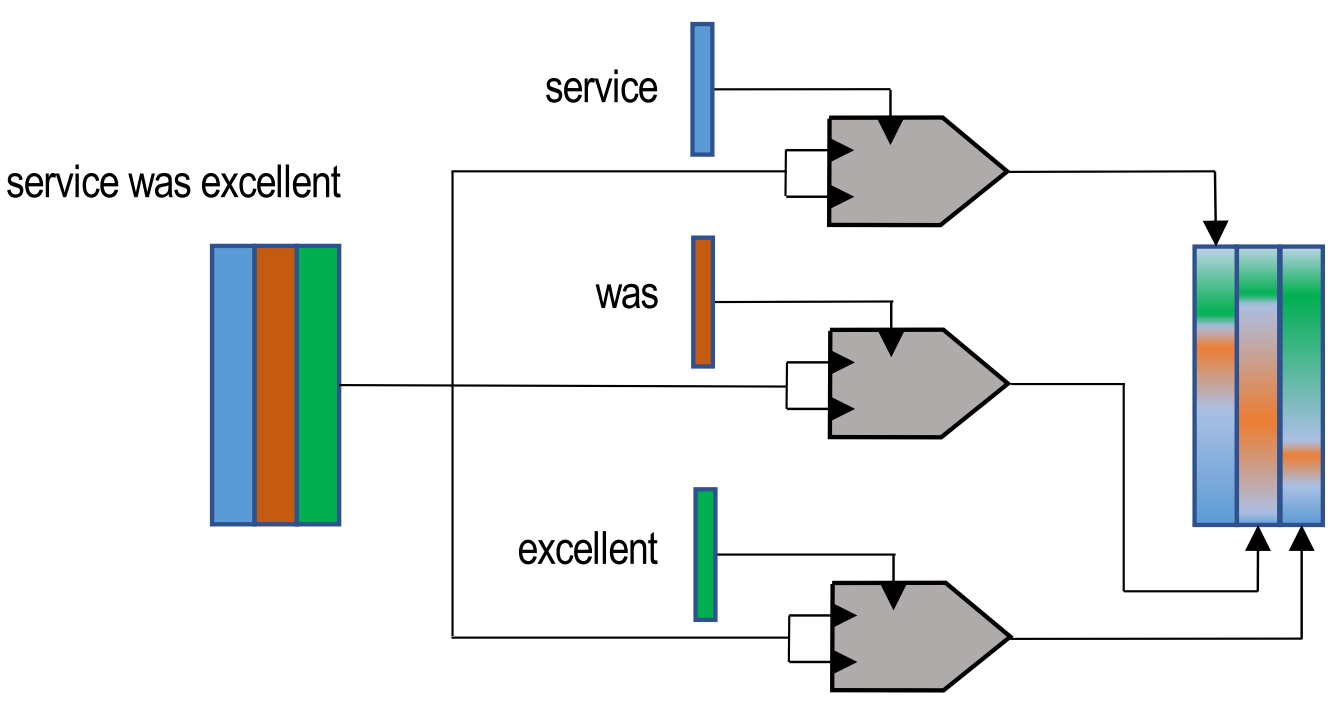

Fig. 5. Example of use of attention in a sequence-to-sequence model.

*2) Hierarchical-Input Architectures:* In some tasks, portions of input data can be meaningfully grouped together into higher level structures, where hierarchical-input attention models can be exploited to subsequently apply multiple attention modules at different levels of the composition, as shown in Fig. 6.

Consider, for instance, data naturally associated with a two-level semantic structure, such as characters (the "micro" elements) forming words (the "macro" elements) or words forming sentences. Attention can be first applied to the representations of micro elements $\boldsymbol{k_i}$, so as to build aggregate representations $\boldsymbol{k_j}$ of the macro elements, such as context vectors. Attention could then be applied again to the sequence of macroelement embeddings, in order to compute an embedding for the whole document $D$. With this model, attention first highlights the most relevant micro elements within each macro element and, then, the most relevant macro elements in the document. For instance, Yang *et al.* [52] applied attention first at word level, for each sentence in turn, to compute sentence embeddings. Then, they applied attention again on the sentence embeddings to obtain a document representation. With reference to the model introduced in Section II, embeddings are computed for each sentence in $D$, and then, all such embeddings are used together as keys $\boldsymbol{K}$ to compute the document-level weights $\boldsymbol{a}$ and eventually $D$'s context vector $\boldsymbol{c}$. The hierarchy can be extended further. For instance, Wu *et al.* [141] added another layer on top, applying attention also at the document level.

If representations for both micro- and macro-level elements are available, one can compute attention on one level and then exploit the result as a key or query to compute attention on the other, yielding two different microrepresentation/macrorepresentation of $D$. In this way, attention enables us to identify the most relevant elements for the task at both levels. The attention-via-attention model by Zhao and Zhang [43] defines a hierarchy with characters at the micro level and words at the macro level. Both characters and words act as keys. Attention is first computed on word embeddings $\boldsymbol{K_W}$, thus obtaining a document representation in the form of a context vector $\boldsymbol{c_W}$, which in turn acts as a query $\boldsymbol{q}$ to guide the application of character-level attention to the keys (character embeddings) $\boldsymbol{K_C}$, yielding a context vector $\boldsymbol{c}$ for $D$.

Ma *et al.* [113] identified a single "target" macro-object $T$ as a set of words, which do not necessarily have to form a sequence in $D$, and then used such a macro-object as keys, $\boldsymbol{K_T}$. The context vector $\boldsymbol{c_T}$ produced by a first application of the attention mechanism on $\boldsymbol{K_T}$ is then used as query $\boldsymbol{q}$ in a second application of the attention mechanism over $D$, with the keys being the document's word embeddings $\boldsymbol{K_W}$.

### B. Compatibility Functions

The compatibility function is a crucial part of the attention architecture because it defines how keys and queries are matched or combined. In our presentation of compatibility functions, we will consider a data model where $\boldsymbol{q}$ and $\boldsymbol{k_i}$ are monodimensional vectors. For example, if $\boldsymbol{K}$ represents a document, each $\boldsymbol{k_i}$ may be the embedding of a sentence, a word, or a character. In such a model, $\boldsymbol{q}$ and $\boldsymbol{k_i}$ may have the same structure and, thus, the same size, although this is not always necessary. However, in some architectures, $\boldsymbol{q}$ can consist of a sequence of vectors or a matrix, a possibility we explore in Section IV-D2.

Some common compatibility functions are listed in Table IV. Two main approaches can be identified. The first one is to match and compare $\boldsymbol{K}$ and $\boldsymbol{q}$. For instance, the idea behind the similarity attention proposed by Graves *et al.* [142] is that the most relevant keys are the most similar to the query. Accordingly, the authors present a model that relies on a similarity function (sim in Table IV) to compute the energy scores. For example, they rely on cosine similarity, a choice that suits cases where the query and the keys share the same semantic representation. A similar idea is followed by the widely used multiplicative or dot attention, where the dot product between $\boldsymbol{q}$ and $\boldsymbol{K}$ is computed. The complexity of this computation is $O(n_k d_k)$. A variation of this model is scaled multiplicative attention, where a scaling factor is introduced to improve performance with large keys [36]. General attention, proposed by Luong *et al.* [29],

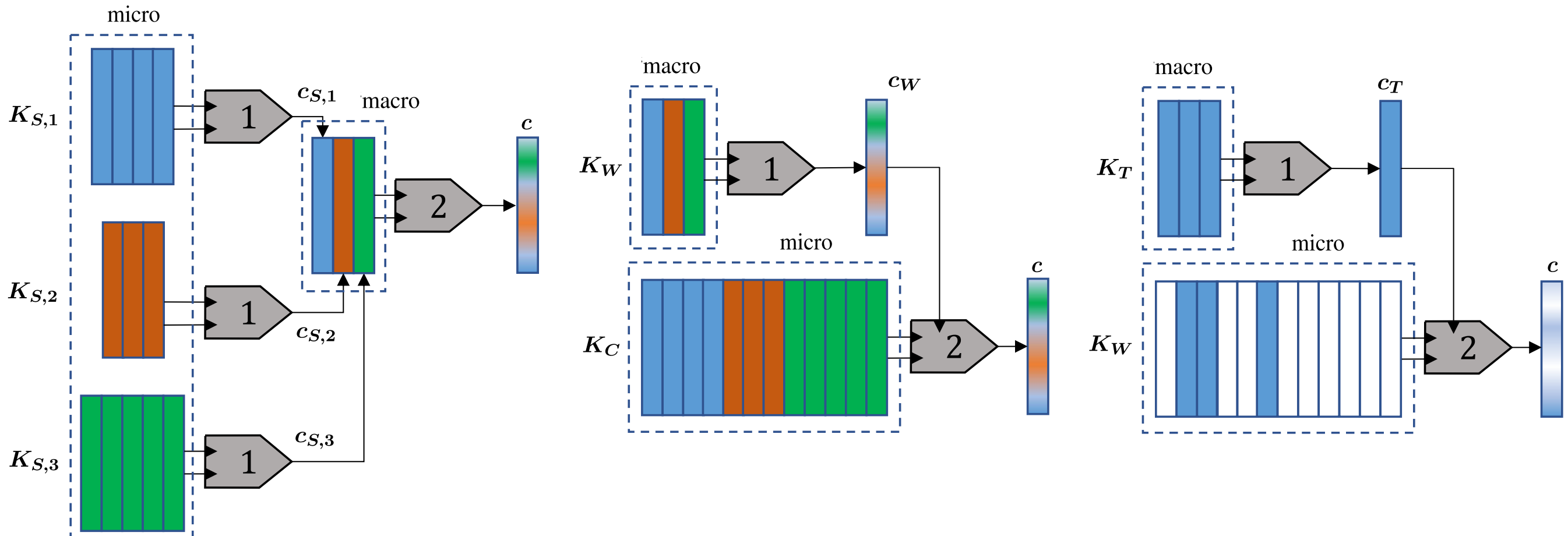

Fig. 6. Hierarchical input attention models defined by Zhao and Zhang [43] (center), Yang *et al.* [52] (left), and Ma *et al.* [113] (right). The number inside the attention shapes indicates the order of application. Different colors highlight different parts of the inputs.

TABLE IV

SUMMARY OF COMPATIBILITY FUNCTIONS FOUND IN THE LITERATURE. $\boldsymbol{W}$, $\boldsymbol{W_0}$, $\boldsymbol{W_1}, \ldots$, AND $\boldsymbol{b}$ ARE LEARNABLE PARAMETERS

| Name | Equation | Ref. |
|---|---|---|
| *similarity* | $f(\boldsymbol{q}, \boldsymbol{K}) = sim(\boldsymbol{q}, \boldsymbol{K})$ | [142] |
| *multiplicative* or *dot* | $f(\boldsymbol{q}, \boldsymbol{K}) = \boldsymbol{q}^\intercal \boldsymbol{K}$ | [29] |
| *scaled multiplicative* | $f(\boldsymbol{q}, \boldsymbol{K}) = \frac{\boldsymbol{q}^\intercal \boldsymbol{K}}{\sqrt{n_k}}$ | [36] |
| *general* or *bilinear* | $f(\boldsymbol{q}, \boldsymbol{K}) = \boldsymbol{q}^\intercal \boldsymbol{W} \boldsymbol{K}$ | [29] |
| *biased general* | $f(\boldsymbol{q}, \boldsymbol{K}) = \boldsymbol{K}^\intercal (\boldsymbol{W}\boldsymbol{q} + \boldsymbol{b})$ | [67] |
| *activated general* | $f(\boldsymbol{q}, \boldsymbol{K}) = act(\boldsymbol{q}^\intercal \boldsymbol{W} \boldsymbol{K} + \boldsymbol{b})$ | [111] |
| *concat* | $f(\boldsymbol{q}, \boldsymbol{K}) = \boldsymbol{w_{imp}}^\intercal act\big(\boldsymbol{W}[\boldsymbol{K}; \boldsymbol{q}] + \boldsymbol{b}\big)$ | [29] |
| *additive* | $f(\boldsymbol{q}, \boldsymbol{K}) = \boldsymbol{w_{imp}}^\intercal act(\boldsymbol{W_1}\boldsymbol{K} + \boldsymbol{W_2}\boldsymbol{q} + \boldsymbol{b})$ | [2] |
| *deep* | $f(\boldsymbol{q}, \boldsymbol{K}) = \boldsymbol{w_{imp}}^\intercal \boldsymbol{E}^{(L-1)} + \boldsymbol{b}^L$<br>$\boldsymbol{E}^{(l)} = act\Big(\boldsymbol{W_l}\boldsymbol{E}^{(l-1)} + \boldsymbol{b}^l\Big)$<br>$\boldsymbol{E}^{(1)} = act\Big(\boldsymbol{W_1}\boldsymbol{K} + \boldsymbol{W_0}\boldsymbol{q} + \boldsymbol{b}^1\Big)$ | [54] |
| *convolution-based* | $f(\boldsymbol{q}, \boldsymbol{K}) = [e_0; \ldots; e_{d_k}]$<br>$e_j = \frac{1}{l}\sum_{i=j-l}^{j} e_{j,i}$<br>$e_{j,i} = act\Big(\boldsymbol{w_{imp}}^\intercal[\boldsymbol{k_i}; \ldots; \boldsymbol{k_{i+l}}] + b\Big)$ | [119] |
| *location-based* | $f(\boldsymbol{q}, \boldsymbol{K}) = f(\boldsymbol{q})$ | [29] |

extends this concept in order to accommodate keys and queries with different representations. To that end, it introduces a learnable matrix parameter $\boldsymbol{W}$. This parameter represents the transformation matrix that maps the query onto the vector space of keys. This transformation increases the complexity of the operation to $O(n_q n_k d_k)$. In what could be called a biased general attention, Sordoni *et al.* [67] introduced a learnable bias, so as to consider some keys as relevant independently of the input. Activated general attention [111] employs a nonlinear activation function. In Table IV, *act* is a placeholder for a nonlinear activation function, such as hyperbolic tangent, tanh, rectifier linear unit, ReLU [143], or scaled exponential linear unit, SELU [144]. These approaches are particularly suitable in tasks where the concept of relevance of a key is known to be closely related to that of similarity to a query element. These include, for instance, tasks where specific keywords can be used as a query, such as abusive speech recognition and sentiment analysis.

A different approach amounts to combining rather than comparing $\boldsymbol{K}$ and $\boldsymbol{q}$, using them together to compute a joint representation, which is then multiplied by an importance vector[3] $\boldsymbol{w_{imp}}$, which has to adhere to the same semantic of the new representation. Such a vector defines, in a way, relevance and could be an additional query element, as offered by Ma *et al.* [113] or a learnable parameter. In that case, we speculate that the analysis of a machine-learned importance vector could provide additional information on the model. One of the simplest models that follows this approach is the concat attention by Luong *et al.* [29], where a joint representation is given by juxtaposing keys and queries. Additive attention works similarly, except that the contribution of $\boldsymbol{q}$ and $\boldsymbol{K}$ can be computed separately. For example, Bahdanau *et al.* [2] precomputed the contribution of $\boldsymbol{K}$ in order to reduce the computational footprint. The complexity of the computation, ignoring the application of the nonlinear function, is thus $O(n_w n_k d_k)$, where $n_w$ indicates the size of $\boldsymbol{w_{imp}}$. Moreover, additive attention in principle could accommodate queries of different size. In additive attention and concat attention, the keys and the queries are fed into a single neural layer. We speak instead of deep attention if multiple layers are employed [54]. Table IV shows a deep attention function with $L$ levels of depth, $1 < l < L$. With an equal number of neurons for all levels and reasonably assuming $L$ to be much smaller than any other parameter, the complexity becomes $O(n_w^L n_k d_k)$. These approaches are especially suitable when a representation of "relevant" elements is unavailable or it is available but encoded in a significantly different way from the way that keys are encoded. This may be the case, for instance, with tasks such as document classification and summarization.

[3]Part of our terminology. As previously noted, wimp is termed context vector by Yang *et al.* [52] and other authors.

A similar rationale lead to convolution-based attention [119]. It draws inspiration from biological models, whereby biological attention is produced through search templates and pattern matching. Here, the learned vector $w_{imp}$ represents a convolutional filter, which embodies a specific relevance template. The filter is used in a pattern-matching process obtained by applying the convolution operation on key subsequences. Applying multiple convolutions with different filters enables a contextual selection of keys that match specific relevance templates, obtaining a specific energy score for each filter. Since each key belongs to multiple subsequences, each key yields multiple energy scores. Such scores are then aggregated in order to obtain a single value per key. Aggregation could be achieved by sum or average. Table IV illustrates convolution-based attention for a single filter of size $l$. The complexity of such an operation is $O(l^2 \; n_k d_k)$. These approaches are especially suitable when a representation of "relevant" elements is unavailable or it is available but encoded in a significantly different way from the way that keys are encoded. This may be the case, for instance, with tasks such as document classification and summarization.

Finally, in some models, the attention distribution ignores $\boldsymbol{K}$ and only depends on $\boldsymbol{q}$. We then speak of location-based attention. The energy associated with each key is thus computed as a function of the key's position, independently of its content [29]. Conversely, self-attention may be computed only based on $\boldsymbol{K}$, without any $\boldsymbol{q}$. The compatibility functions for self-attention, which are a special case of the more general functions, are omitted from Table IV.

For an empirical comparison between some compatibility functions (namely, additive and multiplicative attention) in the domain of machine translation, we suggest the reader to refer to Britz *et al.* [37].

### C. Distribution Functions

Attention distribution maps energy scores to attention weights. The choice of the distribution function depends on the properties that the distribution is required to have—for instance, whether it is required to be a probability distribution, a set of probability scores, or a set of Boolean scores—on the need to enforce sparsity, and on the need to account for the keys' positions.

One possible distribution function $g$ is the logistic sigmoid, as proposed by Kim *et al.* [47]. In this way, each weight $a_i$ is constrained between 0 and 1, thus ensuring that the values $\boldsymbol{V_i}$ and their corresponding weighted counterparts $\boldsymbol{Z_i}$ share the same boundaries. Such weights can thus be interpreted as probabilities that an element is relevant. The same range can also be forced on the context vector's elements $c_i$, by using a softmax function, as it is commonly done. In that case, the attention mechanism is called soft attention. Each attention weight can be interpreted as the probability that the corresponding element is the most relevant.

With sigmoid or softmax alike, all the key/value elements have a relevance, small as it may be. Yet, it can be argued that in some cases, some parts of the input are completely irrelevant, and if they were to be considered, they would likely introduce noise rather than contribute with useful information. In such cases, one could exploit attention distributions that altogether ignore some of the keys, thereby reducing the computational footprint. One option is the sparsemax distribution [91], which truncates to zero the scores under a certain threshold by exploiting the geometric properties of the probability simplex. This approach could be especially useful in those settings where a large number of elements are irrelevant, such as in document summarization or cloze question-answering tasks.

It is also possible to model structural dependences between the outputs. For example, structured attention networks [47] exploit neural conditional random fields to model the (conditional) attention distribution. The attention weights are thus computed as (normalized) marginals from the energy scores, which are treated as potentials.

In some tasks, such as machine translation or image captioning, the relevant features are found in a neighborhood of a certain position. In those cases, it could be helpful to focus the attention only on a specific portion of the input. If the position is known in advance, one can apply a positional mask, by adding or subtracting a given value from the energy scores before the application of the softmax [93]. Since the location may not be known in advance, the hard attention model by Xu *et al.* [16] considers the keys in a dynamically determined location. Such a solution is less expensive at inference time, but it is not differentiable. For that reason, it requires more advanced training techniques, such as reinforcement learning or variance reduction. Local attention [29] extends this idea while preserving differentiability. Guided by the intuition that in machine translation at each time step, only a small segment of the input can be considered relevant, and local attention considers only a small window of the keys at a time. The window has a fixed size, and the attention can be better focused on a precise location by combining the softmax distribution with a Gaussian distribution. The mean of the Gaussian distribution is dynamic, whereas its variance can either be fixed, as done by Luong *et al.* [29] or dynamic, as done by Yang *et al.* [39]. Selective attention [17] follows the same idea; using a grid of Gaussian filters, it considers only a patch of the keys, whose position, size, and resolution depend on dynamic parameters.

Shen *et al.* [94] combined soft and hard attention, by applying the former only on the elements filtered by the latter. More precisely, softmax is only applied among a subset of selected energy scores, whereas for the others, the weight is set to zero. The subset is determined according to a set of random variables, with each variable corresponding to a key. The probability associated with each variable is determined through soft attention applied to the same set of keys. The proper "softness" of the distribution could depend not only on the task but also on the query. Lin *et al.* [44] defined a model whose distribution is controlled by a learnable, adaptive temperature parameter. When a "softer" attention is required, the temperature increases, producing a smoother distribution of weights. The opposite happens when a "harder" attention is needed.

Finally, the concept of locality can also be defined according to semantic rules, rather than the temporal position. This possibility will be further discussed in Section V.

### D. Multiplicity

We shall now present variations of the general unified model where the attention mechanism is extended to accommodate multiple, possibly heterogeneous, inputs or outputs.

*1) Multiple Outputs:* Some applications suggest that the data could, and should, be interpreted in multiple ways. This can be the case when there is ambiguity in the data, stemming, for example, from words having multiple meanings or when addressing a multitask problem. For this reason, models have been defined that jointly compute not only one but multiple attention distributions over the same data. One possibility presented by Lin *et al.* [100] and also exploited by Du *et al.* [145] is to use additive attention (seen in Section IV-B) with an importance matrix, instead of a vector, $\boldsymbol{W_{imp}} \in \mathbb{R}^{n_k \times n_o}$, yielding an energy scores matrix where multiple scores are associated with each key. Such scores can be regarded as different models of relevance for the same values and can be used to create a context matrix $\boldsymbol{C} \in \mathbb{R}^{n_v \times n_o}$. Such embeddings can be concatenated together, creating a richer and more expressive representation of the values. Regularization penalties can be applied so as to enforce the differentiation between models of relevance (for example, the Frobenius norm). In multidimensional attention [93], where the importance matrix is a square matrix, attention can be computed featurewise. To that end, each weight $a_{i,j}$ is paired with the $j$th feature of the $i$th value $v_{i,j}$, and a featurewise product yields the new value $z_i$. Convolution-based attention [119] always produces multiple energy scores distributions according to the number of convolutional filters and the size of those filters. Another possibility explored by Vaswani *et al.* [36] is multihead attention. There, multiple linear projections of all the inputs ($\boldsymbol{K}$, $\boldsymbol{V}$, and $\boldsymbol{q}$) are performed according to learnable parameters, and multiple attention functions are computed in parallel. Usually, the processed context vectors are then merged together into a single embedding. A suitable regularization term is sometimes imposed so as to guarantee sufficient dissimilarity between attention elements. Li *et al.* [34] proposed three possibilities: regularization on the subspaces (the linear projections of $\boldsymbol{V}$), the attended positions (the sets of weights), or on the outputs (the context vectors). Multihead attention can be especially helpful when combined with nonsoft attention distribution since different heads can capture local and global contexts at the same time [39]. Finally, labelwise attention [126] computes a separate attention distribution for each class. This may improve the performance as well as lead to a better interpretation of the data because it could help isolate data points that better describe each class. These techniques are not mutually exclusive. For example, multihead and multidimensional attention can be combined with one another [95].

*2) Multiple Inputs: Coattention:* Some architectures consider the query to be a sequence of $d_q$ multidimensional elements, represented by a matrix $\boldsymbol{Q} \in \mathbb{R}^{n_q \times d_q}$, rather than by a plain vector. Examples of this setup are common in architectures designed for tasks where the query is a sentence, as in question answering, or a set of keywords, as in abusive speech detection. In those cases, it could be useful to find the most relevant query elements according to the task and the keys. A straightforward way of doing that would be to apply the attention mechanism to the query elements, thus treating $\boldsymbol{Q}$ as keys and each $\boldsymbol{k_i}$ as query, yielding two independent representations for $\boldsymbol{K}$ and $\boldsymbol{Q}$. However, in that way, we would miss the information contained in the interactions between elements of $\boldsymbol{K}$ and $\boldsymbol{Q}$. Alternatively, one could apply attention jointly on $\boldsymbol{K}$ and $\boldsymbol{Q}$, which become the "inputs" of a coattention architecture [80]. Coattention models can be coarse-grained or fine-grained [112]. Coarse-grained models compute attention on each input, using an embedding of the other input as a query. Fine-grained models consider each element of an input with respect to each element of the other input. Furthermore, coattention can be performed sequentially or in parallel. In parallel models, the procedures to compute attention on $\boldsymbol{K}$ and $\boldsymbol{Q}$ symmetric, and thus, the two inputs are treated identically.

*a) Coarse-grained coattention:* Coarse-grained models use a compact representation of one input to compute attention on the other. In such models, the role of the inputs as keys and queries is no longer focal, and thus, a compact representation of $\boldsymbol{K}$ may play as a query in parts of the architecture and vice versa. A sequential coarse-grained model proposed by Lu *et al.* [80] is alternating coattention, as shown in Fig. 7 (left), whereby attention is computed three times to obtain embeddings for $\boldsymbol{K}$ and $\boldsymbol{Q}$. First, self-attention is computed on $\boldsymbol{Q}$. The resulting context vector is then used as a query to perform attention on $\boldsymbol{K}$. The result is another context vector $\boldsymbol{C_K}$, which is further used as a query as attention is again applied to $\boldsymbol{Q}$. This produces a final context vector, $\boldsymbol{C_Q}$. The architecture proposed by Sordoni *et al.* [67] can also be described using this model with a few adaptations. In particular, Sordoni *et al.* [67] omitted the last step and factor in an additional query element $\boldsymbol{q}$ in the first two attention steps. An almost identical sequential architecture is used by Zhang *et al.* [86], who use $\boldsymbol{q}$ only in the first attention step. A parallel coarse-grained model is shown in Fig. 7 (right). In such a model, proposed by Ma *et al.* [111], an average ($avg$) is initially computed on each input and then used as a query in the application of attention to generate the embedding of the other input. Sequential coattention offers a more elaborate computation of the final results, potentially allowing to discard all the irrelevant elements of $Q$ and $K$, at the cost of a greater computational footprint. Parallel coattention can be optimized for better performance, at the expense of a "simpler" elaboration of the outputs. It is worthwhile noticing that the averaging step in Ma *et al.* [111]'s model could be replaced by self-attention, in order to filter out irrelevant elements from $Q$ and $K$ at an early stage.

*b) Fine-grained coattention:* In fine-grained coattention models, the relevance (energy scores) associated with each key/query element pair $\langle \boldsymbol{k_i}/\boldsymbol{q_j} \rangle$ is represented by the elements $E_{j,i}$ of a coattention matrix $\boldsymbol{E} \in \mathbb{R}^{d_q \times d_k}$ computed by a cocompatibility function.

Cocompatibility functions can be straightforward adaptations of any of the compatibility functions listed in Table IV. Alternatively, new functions can be defined. For example, Fan *et al.* [112] defined cocompatibility as a linear transformation of the concatenation between the elements and their product [see (12)]. In decomposable attention [89], the inputs are fed into neural networks, whose outputs are then multiplied [see (13)]. Delaying the product to after the processing by the neural networks reduces the number of inputs of such networks, yielding a reduction in the computational footprint. An alternative proposed by Tay *et al.* [75] exploits the Hermitian inner product. The elements of $\boldsymbol{K}$ and $\boldsymbol{Q}$ are first projected to a complex domain, then, the Hermitian product between them is computed, and finally, only the real part of the result is kept. Being the Hermitian product noncommutative, $\boldsymbol{E}$ will depend on the roles played by the inputs as keys and queries

$$E_{j,i} = \boldsymbol{W}([\boldsymbol{k_i}; \boldsymbol{q_j}; \boldsymbol{k_i}\boldsymbol{q_j}]) \tag{12}$$

$$E_{j,i} = (\mathrm{act}(\boldsymbol{W_1}\boldsymbol{Q} + \boldsymbol{b_1}))^{\mathsf{T}}(\mathrm{act}(\boldsymbol{W_2}\boldsymbol{K} + \boldsymbol{b_2})). \tag{13}$$

Because $E_{j,i}$ represent energy scores associated with $\langle \boldsymbol{k_i}/\boldsymbol{q_j} \rangle$ pairs, computing the relevance of $\boldsymbol{K}$ with respect to specific query elements, or, similarly, the relevance of $\boldsymbol{Q}$ with respect to specific key elements, requires extracting information from $\boldsymbol{E}$ using what we call an aggregation function. The output of such a function is a pair $\boldsymbol{a_K}/\boldsymbol{a_Q}$ of weight vectors.

The commonest aggregation functions are listed in Table V. A simple idea is the attention pooling parallel model adopted by dos Santos *et al.* [69]. It amounts to considering the highest score in each row or column of $\boldsymbol{E}$. By attention pooling, a key $\boldsymbol{k_i}$ will be attributed a high attention weight if and only if it has a high coattention score with respect to at least one query element $\boldsymbol{q_j}$. Key attention scores are obtained through rowwise max pooling, whereas query attention scores are obtained through columnwise max pooling, as shown in Fig. 8 (left). Only considering the highest attention scores may be regarded as a conservative approach. Indeed, low-energy scores can only be obtained by keys whose coattention scores are all low and thus likely to be irrelevant to all query elements. Furthermore, the keys that are only relevant to a single query element may obtain the same or even higher energy score than keys that are relevant to all the query elements. The same reasoning applies to query attention scores. This is a suitable approach, for example, in tasks where the queries are keywords in disjunction, such as in abusive speech recognition. Conversely, the approaches follow to compute the energy score of an element by accounting for all the related attention scores, at the cost of a heavier computational footprint.

TABLE V

AGGREGATION FUNCTIONS. IN MOST CASES, $\boldsymbol{a_K}$ AND $\boldsymbol{a_Q}$ ARE OBTAINED BY APPLYING A DISTRIBUTION FUNCTION, SUCH AS THOSE SEEN IN SECTION IV-C, TO $\boldsymbol{e_K}$ AND $\boldsymbol{e_Q}$, AND ARE THUS OMITTED FROM THIS TABLE IN THE INTEREST OF BREVITY. AS CUSTOMARY, *Act* IS A PLACEHOLDER FOR A GENERIC NONLINEAR ACTIVATION FUNCTION, WHEREAS *Dist* INDICATES A DISTRIBUTION FUNCTION SUCH AS SOFTMAX

| **Name** | **Equations** | **Ref.** |
|---|---|---|
| *pooling* | $e_{Ki} = \max_{1 \le j \le d_q} (\boldsymbol{E_{j,i}})$<br>$e_{Qj} = \max_{1 \le i \le d_k} (\boldsymbol{E_{j,i}})$ | [69] |
| *perceptron* | $\boldsymbol{e_K} = \boldsymbol{W_3}^{\mathsf{T}} act(\boldsymbol{W_1}\boldsymbol{K} + \boldsymbol{W_2}\boldsymbol{Q}\boldsymbol{E})$<br>$\boldsymbol{e_Q} = \boldsymbol{W_4}^{\mathsf{T}} act(\boldsymbol{W_2}\boldsymbol{K} + \boldsymbol{W_1}\boldsymbol{Q}\boldsymbol{E}^{\mathsf{T}})$ | [80] |
| *linear transformation* | $\boldsymbol{e_K} = \boldsymbol{W_1}\boldsymbol{E}$<br>$\boldsymbol{e_Q} = \boldsymbol{W_2}\boldsymbol{E}$ | [127] |
| *attention over attention* | $\boldsymbol{a_K} = \boldsymbol{M_1} \cdot \boldsymbol{a_Q}$<br>$\boldsymbol{a_Q} = \underset{1 \le i \le d_k}{average}(\boldsymbol{M_{2,i}})$<br>$\boldsymbol{M_{2,i}} = \underset{1 \le j \le d_q}{dist}(\boldsymbol{E_{j,i}})$<br>$\boldsymbol{M_{1,j}} = \underset{1 \le i \le d_k}{dist}(\boldsymbol{E_{j,i}})$ | [70] |
| *perceptron with nested attention* | $\boldsymbol{e_K} = \boldsymbol{W_3}^{\mathsf{T}} act\big(\boldsymbol{W_1}\boldsymbol{K} + (\boldsymbol{W_2}\boldsymbol{Q}^{\mathsf{T}})\boldsymbol{M_2}\big)$<br>$\boldsymbol{e_Q} = \boldsymbol{W_4}^{\mathsf{T}} act\big(\boldsymbol{W_2}\boldsymbol{Q} + (\boldsymbol{W_1}\boldsymbol{K}^{\mathsf{T}})\boldsymbol{M_1}^{\mathsf{T}}\big)$<br>$\boldsymbol{M_{2,i}} = \underset{1 \le j \le d_q}{dist}(\boldsymbol{E_{j,i}})$<br>$\boldsymbol{M_{1,j}} = \underset{1 \le i \le d_k}{dist}(\boldsymbol{E_{j,i}})$ | [88] |

Lu *et al.* [80] used a multilayer perceptron in order to learn the mappings from $\boldsymbol{E}$ to $\boldsymbol{e_K}$ and $\boldsymbol{e_Q}$. In [127], the computation is even simpler since the final energy scores are a linear transformation of $\boldsymbol{E}$. Cui *et al.* [70] instead applied the nested model shown in Fig. 8 (right). First, two matrices $\boldsymbol{M_1}$ and $\boldsymbol{M_2}$ are computed by separately applying a rowwise and a columnwise softmax on $\boldsymbol{E}$. The idea is that each row of $\boldsymbol{M_1}$ represents the attention distribution over the document according to a specific query element—and it could already be used as such. Then, a rowwise average over $\boldsymbol{M_2}$ is computed so as to produce an attention distribution $\boldsymbol{a_Q}$ over query elements. Finally, a weighted sum of $\boldsymbol{M_1}$ according to the relevance of query elements is computed through the dot product between $\boldsymbol{M_1}$ and $\boldsymbol{a_Q}$, obtaining the document's attention distribution over the keys, $\boldsymbol{a_K}$. An alternative nested attention model is proposed by Nie *et al.* [88], whereby $\boldsymbol{M_1}$ and $\boldsymbol{M_2}$ are fed to a multilayer perceptron, as is done by Lu *et al.* [80]. Further improvements may be obtained by combining the results of multiple coattention models. Fan *et al.* [112], for instance, computed coarse-grained and fine-grained attention in parallel and combined the results into a single embedding.

## V. COMBINING ATTENTION AND KNOWLEDGE

According to LeCun *et al.* [146], a major open challenge in artificial intelligence (AI) is combining connectionist (or subsymbolic) models, such as deep networks, with approaches based on symbolic knowledge representation, in order to perform complex reasoning tasks. Throughout the last decade, filling the gap between these two families of AI methodologies has represented a major research avenue. Popular approaches include statistical relational learning [147], neural-symbolic learning [148], and the application of various deep learning architectures [149], such as memory networks [59], neural Turing machines [142], and several others.

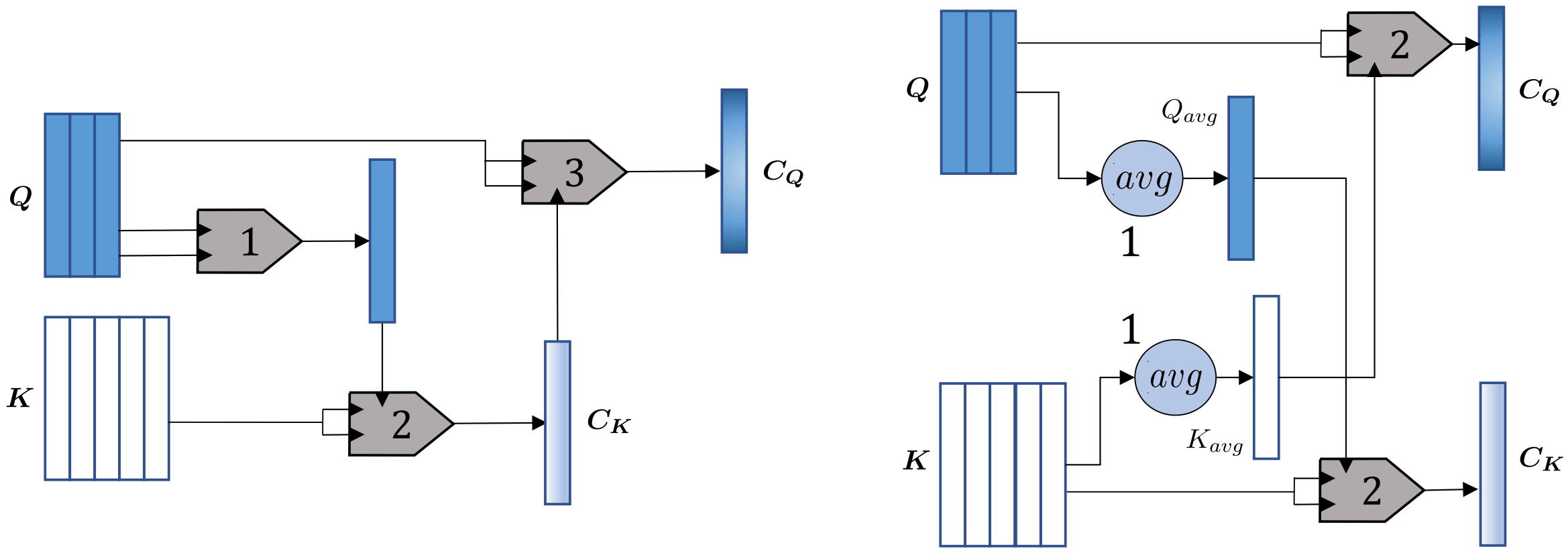

Fig. 7. Coarse-grained coattention by Lu *et al.* [80] (left) and Ma *et al.* [111] (right). The number inside the attention shapes (and besides the average operator) indicates the order in which they are applied. Different colors highlight different inputs.

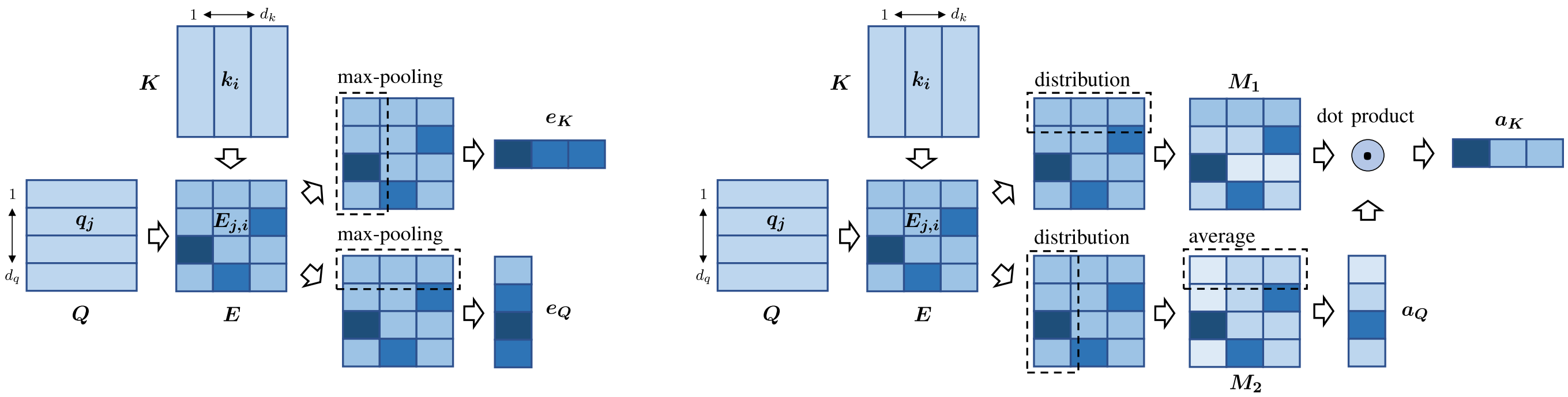

Fig. 8. Fine-grained coattention models presented by dos Santos *et al.* [69] (left) and by Cui *et al.* [70] (right). Dashed lines show how max-pooling/distribution functions are applied (columnwise or rowwise).

From this perspective, attention can be seen both as an attempt to improve the interpretability of neural networks and as an opportunity to plug external knowledge into them. As a matter of fact, since the weights assigned by attention represent the relevance of the input with respect to the given task, in some contexts, it could be possible to exploit this information to isolate the most significant features that allow the deep network to make its predictions. On the other hand, any prior knowledge regarding the data, the domain, or the specific task, whenever available, could be exploited to generate information about the desired attention distribution, which could be encoded within the neural architecture.

In this section, we overview different techniques that can be used to inject this kind of knowledge in a neural network. We leave to Section VI further discussions on the open challenges regarding the combination of knowledge and attention.

### A. Supervised Attention

In most of the works we surveyed, the attention model is trained with the rest of the neural architecture to perform a specific task. Although trained alongside a supervised procedure, the attention model per se is trained in an unsupervised fashion[4] to select useful information for the rest of the architecture. Nevertheless, in some cases, knowledge about the desired weight distribution could be available. Whether it is present in the data as a label or it is obtained as additional information through external tools, it can be exploited to perform a supervised training of the attention model.

*1) Preliminary Training:* One possibility is to use an external classifier. The weights learned by such a classifier are subsequently plugged into the attention model of a different architecture. We name this procedure as preliminary training. For example, Zhang *et al.* [53] first trained an attention model to represent the probability that a sentence contains relevant information. The relevance of a sentence is given by rationales [150], which are snippets of text that supports the corresponding document categorizations. In work by Long *et al.* [118], a model is preliminarily trained on eye-tracking data sets to estimate the reading time of words. Then, the reading time predicted by the model is used as an energy score in a neural model for sentiment analysis.

*2) Auxiliary Training:* Another possibility is to train the attention model without preliminary training, but by treating attention learning as an auxiliary task that is performed jointly with the main task. This procedure has led to good results in many scenarios, including machine translation [30], [35], visual question answering [137], and domain classification for natural language understanding [138].

[4]Meaning that there is no target distribution for the attention model.

In some cases, this mechanism can also be exploited to have attention model-specific features. For example, since the linguistic information is useful for semantic role labeling, attention can be trained in a multitask setting to represent the syntactic structure of a sentence. Indeed, in LISA [97], a multilayer multiheaded architecture for semantic role labeling, one of the attention heads is trained to perform dependence parsing as an auxiliary task.

*3) Transfer Learning:* Furthermore, it is possible to perform transfer learning across different domains [1] or tasks [115]. By performing a preliminary training of an attentive architecture on a source domain to perform a source task, a mapping between the inputs and the distribution of weights will be learned. Then, when another attentive architecture is trained on the target domain to perform the target task, the pretrained model can be exploited. Indeed, the desired distribution can be obtained through the first architecture. Attention learning can, therefore, be treated as an auxiliary task as in the previously mentioned cases. The difference is that the distribution of the pretrained model is used as ground truth, instead of using data labels.

### *B. Attention Tracking*

When attention is applied multiple times on the same data, as in sequence-to-sequence models, a useful piece of information could be how much relevance has been given to the input along different model iterations. Indeed, one may need to keep track of the weights that the attention model assigns to each input. For example, in machine translation, it is desirable to ensure that all the words of the original phrase are considered. One possibility to maintain this information is to use a suitable structure and provide it as an additional input to the attention model. Tu *et al.* [33] exploited a piece of symbolic information called coverage to keep track of the weight associated with the inputs. Every time attention is computed, such information is fed to the attention model as a query element, and it is updated according to the output of the attention itself. In [31], the representation is enhanced by making use of a subsymbolic representation for the coverage.

### *C. Modeling the Distribution Function by Exploiting Prior Knowledge*

Another component of the attention model where prior knowledge can be exploited is the distribution function. For example, constraints can be applied to the computation of the new weights to enforce the boundaries on the weights assigned to the inputs. In [46] and [51], the coverage information is exploited by a constrained distribution function, regulating the amount of attention that the same word receives over time.

Prior knowledge could also be exploited to define or to infer a distance between the elements in the domain. Such domain-specific distance could then be considered in any position-based distribution function, instead of the positional distance. An example of distance could be derived by the syntactical information. Chen *et al.* [40] and He *et al.* [109] used distribution functions that consider the distance between two words along the dependence graph of a sentence.

## VI. Challenges and Future Directions

In this section, we discuss open challenges and possible applications of the attention mechanism in the analysis of neural networks, as a support of the training process and as an enabling tool for the integration of symbolic representations within neural architectures.

### *A. Attention for Deep Networks Investigation*

Whether attention may or may not be considered as a mean to explain neural networks is currently an open debate. Some recent studies [10], [11] suggest that attention cannot be considered a reliable mean to explain or even interpret neural networks. Nonetheless, other works [6]–[9] advocate the use of attention weights as an analytic tool. Specifically, Jain and Wallace [10] proved that attention is not consistent with other explainability metrics and that it is easy to create local adversarial distributions (distributions that are similar to the trained model but produce a different outcome). Wiegreffe and Pinter [9] pushed the discussion further, providing experiments that demonstrate that creating an effective global adversarial attention model is much more difficult than creating a local one and that attention weights may contain information regarding feature importance. Their conclusion is that attention may indeed provide an explanation of a model, if by explanation, we mean a plausible, but not necessarily faithful, reconstruction of the decision-making process, as suggested by Rudin [151] and Riedl [152].

In the context of a multilayer neural architecture, it is fair to assume that the deepest levels will compute the most abstract features [146], [153]. Therefore, the application of attention to deep networks could enable the selection of higher level features, thus providing hints to understand which complex features are relevant for a given task. Following this line of inquiry in the computer vision domain, Zhang *et al.* [18] showed that the application of attention to middle-to-high level feature sets leads to better performance in image generation. The visualization of the self-attention weights has revealed that higher weights are not attributed to proximate image regions, but rather to those regions whose color or texture is most similar to that of the query image point. Moreover, the spatial distribution does not follow a specific pattern, but instead, it changes, modeling a region that corresponds to the object depicted in the image. Identifying abstract features in an input text might be less immediate than in an image, where the analysis process is greatly aided by visual intuition. Yet, it may be interesting to test the effects of the application of attention at different levels and to assess whether its weights correspond to specific high-level features. For example, Vaswani *et al.* [36] analyzed the possible relation between attention weights and syntactic predictions, Voita *et al.* [49] did the same with anaphora resolutions, and Clark *et al.* [6] investigated the correlation with many linguistic features. Voita *et al.* [50] analyzed the behavior of the heads of a multihead model, discovering that different heads develop different behaviors, which can be related to specific position or specific syntactical element. Yang *et al.* [39] seemed to confirm that the deeper levels of neural architectures capture nonlocal aspects of the textual input. They studied the application of locality at different depths of an attentive deep architecture and showed that its introduction is especially beneficial when it is applied to the layers that are closer to the inputs. Moreover, when the application of locality is based on a variable-size window, higher layers tend to have a broader window.

A popular way of investigating whether an architecture has learned high-level features amounts to using the same architecture to perform other tasks, as it happens with transfer learning. This setting has been adopted outside the context of attention, for example, by Shi *et al.* [154], who perform syntactic predictions by using the hidden representations learned with machine translation. In a similar way, attention weights could be used as input features in a different model, so as to assess whether they can select relevant information for a different learning task. This is what happens, for example, in attention distillation, where a student network is trained to penalize the most confusing features according to a teacher network, producing an efficient and robust model in the task of machine reading comprehension [155]. Similarly, in a transfer learning setting, attentional heterogeneous transfer [156] has been exploited in heterolingual text classification to selectively filter input features from heterogeneous sources.

### *B. Attention for Outlier Detection and Sample Weighing*

Another possible use of attention may be for outlier detection. In tasks such as classification or the creation of a representative embedding of a specific class, attention could be applied over all the samples belonging to that task. In doing so, the samples associated with small weights could be regarded as outliers with respect to their class. The same principle could be potentially applied to each data point in a training set, independently of its class. The computation of a weight for each sample could be interpreted as assessing the relevance of that specific data point for a specific task. In principle, assigning such samples a low weight and excluding them from the learning could improve a model's robustness to noisy input. Moreover, a dynamic computation of these weights during training would result in a dynamic selection of different training data in different training phases. While attention-less adaptive data selection strategies have already proven to be useful for efficiently obtaining more effective models [117], to the best of our knowledge, no attention-based approach has been experimented to date.

### *C. Attention Analysis for Model Evaluation*

The impact of attention is greatest when all the irrelevant elements are excluded from the input sequence, and the importance of the relevant elements is properly balanced. A seemingly uniform distribution of the attention weights could be interpreted as a sign that the attention model has been unable to identify the more useful elements. This, in turn, may be due to the data that do not contain useful information for the task at hand or it may be ascribed to the poor ability of the model to discriminate information. Nevertheless, the attention model would be unable to find relevant information in the specific input sequence, which may lead to errors. The analysis of the distribution of the attention weights may, therefore, be a tool for measuring an architecture's confidence in performing a task on a given input. We speculate that a high entropy in the distribution or the presence of weights above a certain threshold may be correlated with a higher probability of success of the neural model. These may, therefore, be used as indicators, to assess the uncertainty of the architecture, as well as to improve its interpretability. Clearly, this information would be useful to the user, to better understand the model and the data, but it may also be exploited by more complex systems. For example, Heo *et al.* [157] proposed to exploit the uncertainty of their stochastic predictive model to avoid making risky predictions in healthcare tasks.

In the context of an architecture that relies on multiple strategies to perform its task, such as a hybrid model that relies on both symbolic and subsymbolic information, the uncertainty of the neural model can be used as a parameter in the merging strategy. Other contexts in which this information may be relevant are multitask learning and reinforcement learning. Examples of exploitation of the uncertainty of the model, although in contexts other than attention and NLP, can be found in works by Poggi and Mattoccia [158], Kendall *et al.* [159], and Blundell *et al.* [160].

### *D. Unsupervised Learning With Attention*

To properly exploit unsupervised learning is widely recognized as one of the most important long-term challenges of AI [146]. As already mentioned in Section V, attention is typically trained in a supervised architecture, although without a direct supervision on the attention weights. Nevertheless, a few works have recently attempted to exploit attention within purely unsupervised models. We believe this to be a promising research direction, as the learning process of humans is indeed largely unsupervised.

For example, in work by He *et al.* [161], attention is exploited in a model for aspect extraction in sentiment analysis, with the aim to remove words that are irrelevant for the sentiment and to ensure more coherence of the predicted aspects. In work by Zhang and Wu [162], attention is used within autoencoders in a question-retrieval task. The main idea is to generate semantic representations of questions, and self-attention is exploited during the encoding and decoding phases, with the objective to reconstruct the input sequences, as in traditional autoencoders. Following a similar idea, Zhang *et al.* [163] exploited bidimensional attention-based recursive autoencoders for bilingual phrase embeddings, whereas Tian and Fang [164] exploited attentive autoencoders to build sentence representations and performed topic modeling on short texts. Yang *et al.* [165] instead adopted an attention-driven approach to unsupervised sentiment modification in order to explicitly separate sentiment words from content words.

In computer vision, attention alignment has been proposed for unsupervised domain adaptation, with the aim to align the attention patterns of networks trained on the source and target domain, respectively [166]. We believe that this could be an interesting scenario also for NLP.

### *E. Neural-Symbolic Learning and Reasoning*

Recently, attention mechanisms started to be integrated within some neural-symbolic models, whose application to NLP scenarios is still at an early stage. For instance, in the context of neural logic programming [167], they have been exploited for reasoning over knowledge graphs, in order to combine parameter and structure learning of first-order logic rules. They have also been used in logic attention networks [168] to aggregate information coming from graph neighbors with both rule- and network-based attention weights. Moreover, prior knowledge has also been exploited by Shen *et al.* [169] to enable the attention mechanism to learn the knowledge representation of entities for ranking question–answer pairs.

Neural architectures exploiting attention performed well also in reasoning tasks that are also addressed with symbolic approaches, such as textual entailment [99]. For instance, Hudson and Manning [170] recently proposed a new architecture for complex reasoning problems, with NLP usually being one of the target sub-tasks, as in the case of visual question answering. In such an architecture, attention is used within several parts of the model, for example, over question words or to capture long-range dependences with self-attention.

An attempt to introduce constraints in the form of logical statements within neural networks has been proposed in [171] where rules governing attention are used to enforce word alignment in tasks, such as machine comprehension and natural language inference.

## VII. Conclusion

Attention models have become ubiquitous in NLP applications. Attention can be applied to different input parts, different representations of the same data, or different features, to obtain a compact representation of the data as well as to highlight the relevant information. The selection is performed through a distribution function, which may consider locality in different dimensions, such as space, time, or even semantics. Attention can be used to compare the input data with a query element based on measures of similarity or significance. It can also autonomously learn what is to be considered relevant, by creating a representation encoding what the important data should be similar to. Integrating attention in neural architectures may thus yield a significant performance gain. Moreover, attention can be used as a tool for investigating the behavior of the network.

In this article, we have introduced a taxonomy of attention models, which enabled us to systematically chart a vast portion of the approaches in the literature and compare them to one another. To the best of our knowledge, this is the first systematic, comprehensive taxonomy of attention models for NLP.

We have also discussed the possible role of attention in addressing fundamental AI challenges. In particular, we have shown how attention can be instrumental in injecting knowledge into the neural model, so as to represent specific features, or to exploit previously acquired knowledge, as in transfer learning settings. We speculate that this could pave the way to new challenging research avenues, where attention could be exploited to enforce the combination of subsymbolic models with symbolic knowledge representations to perform reasoning tasks or to address natural language understanding. Recent results also suggest that attention could be a key ingredient of unsupervised learning architectures, by guiding and focusing the training process where no supervision is given in advance.

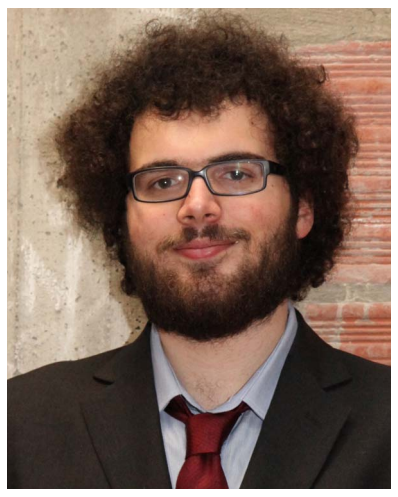

**Andrea Galassi** received the master's degree in computer engineering from the Department of Computer Science and Engineering, University of Bologna, Bologna, Italy, in 2017, where he is currently pursuing the Ph.D. degree.

His research activity concerns artificial intelligence and machine learning, focusing on argumentation mining and related natural language processing (NLP) tasks. Other research interests involve deep learning applications to games and constraint satisfaction problems (CSPs).

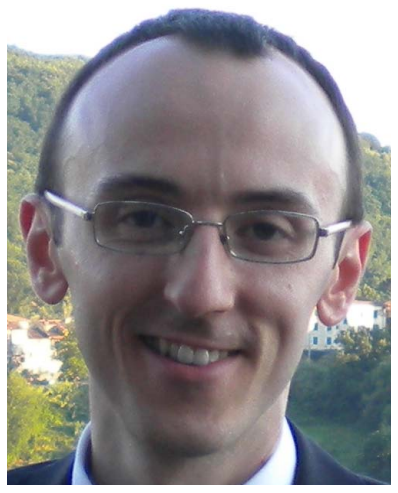

**Marco Lippi** held positions at the University of Florence, Florence, Italy, University of Siena, Siena, Italy, and the University of Bologna, Bologna, Italy. He was a Visiting Scholar with Université Pierre et Marie Curie (UPMC), Paris, France. He is currently an Associate Professor with the Department of Sciences and Methods for Engineering, University of Modena and Reggio Emilia, Modena, Italy. His work focuses on machine learning and artificial intelligence, with applications to several areas, including argumentation mining, legal informatics, and medicine.

Prof. Lippi received the "E. Caianiello" Prize for the Best Italian Ph.D. Thesis in the field of neural networks in 2012.

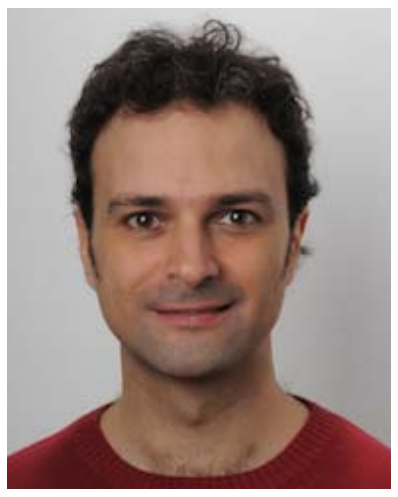

**Paolo Torroni** has been an Associate Professor with the University of Bologna, Bologna, Italy, since 2015. He has edited over 20 books and special issues and authored over 150 articles in computational logics, multiagent systems, argumentation, and natural language processing. His main research interest includes artificial intelligence.

Prof. Torroni serves as an Associate Editor for *Fundamenta Informaticae* and *Intelligenza Artificiale*.